\documentclass[10pt,twocolumn,letterpaper]{article}

\usepackage{cvpr}
\usepackage{times}
\usepackage{epsfig}
\usepackage{graphicx}
\usepackage{amsmath}
\usepackage{amssymb}
\usepackage{gensymb}

\usepackage[keeplastbox]{flushend}

\usepackage{pifont}
\newcommand{\cmark}{\ding{51}}%
\newcommand{\xmark}{\ding{55}}%

\makeatletter
\newcommand{\printfnsymbol}[1]{%
  \textsuperscript{\@fnsymbol{#1}}%
}
\makeatother

\usepackage[breaklinks=true,bookmarks=false]{hyperref}

\cvprfinalcopy 

\def\cvprPaperID{****} 

\begin{document}

\title{Leveraging Temporal Information for 3D Detection and Domain Adaptation}

\author{Cunjun Yu\thanks{equal contribution} \qquad 
Zhongang Cai\printfnsymbol{1} \qquad
Daxuan Ren \qquad
Haiyu Zhao \\
Innova\qquad\\
\tt\small{yucunjun2014@yahoo.com \qquad cai.zhongang@gmail.com}
}

\maketitle

\begin{abstract}
    Ever since the prevalent use of the LiDARs in autonomous driving, tremendous improvements have been made to the learning on the point clouds. However, recent progress largely focuses on detecting objects in a single 360-degree sweep, without extensively exploring the temporal information. In this report, we describe a simple way to pass such information in the learning pipeline by adding timestamps to the point clouds, which shows consistent improvements across all three classes.
\end{abstract}

\begin{figure*}[b]
\centering
  \includegraphics[width=\textwidth]{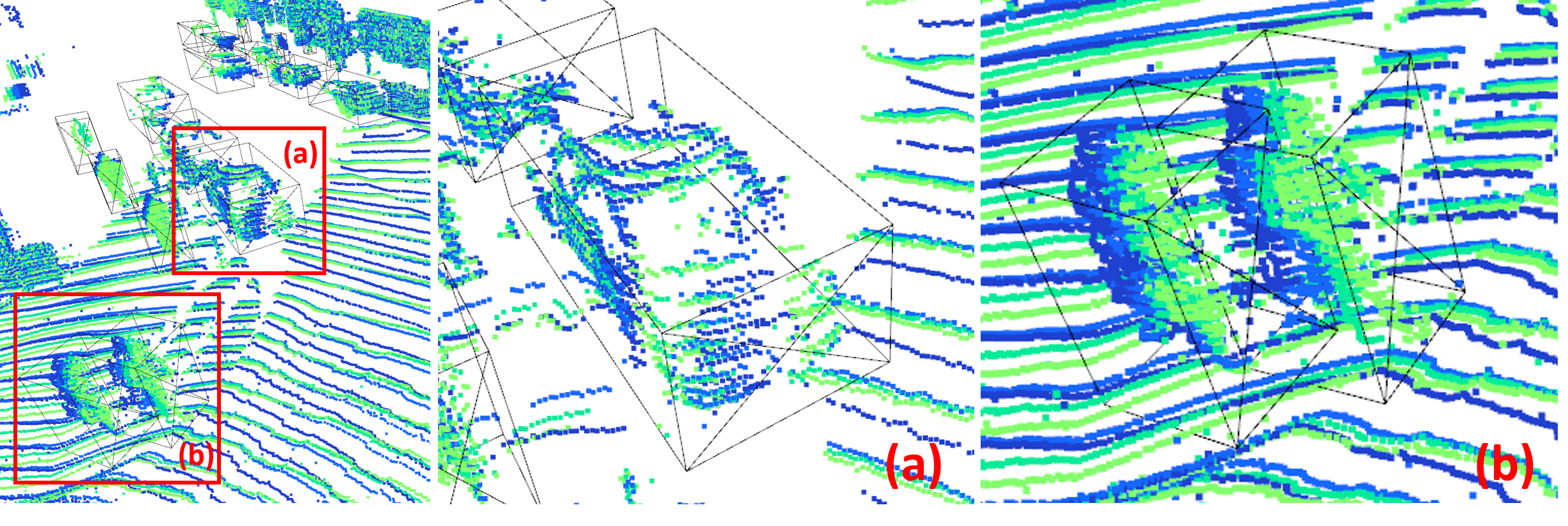}
  \caption{Concatenation of four consecutive frames of point clouds. The current frame is rendered light green and the older the point cloud, the darker the color. The ground truth bounding boxes are drawn in black lines. The heading is indicated by the side with a cross. (a) Each of the point clouds gives a partial description of the car, but when to put together, they form a more complete and denser point cloud. (b) These two moving pedestrians leave ``trails" behind them, which serves as a perfect cue for the heading prediction}
\label{fig:temporal}
\end{figure*}

\section{Introduction}
The point clouds describe the environment by 3D points, eliminating many perspective-related ambiguities that are persistent in the conventional camera-based vision systems. Hence, LiDARs, which output accurate point clouds even at a long range, have become a critical part of the perception pipeline in the realm of autonomous driving. The advent of the KITTI dataset\cite{geiger2013vision} leads to a flourishing development on 3D detection algorithms. However, KITTI provides its data in a discrete, non-sequential form. Hence, many recent works \cite{yan2018second, lang2019pointpillars, shi2019pointrcnn, shi2020points} benchmarked on KITTI focus on a single 360-degree LiDAR sweep for object detection. 

However, the sparsity of points far away from the sensor makes it difficult to detect an object with only a few points in a single sweep. Some pioneering works have attempted to incorporate temporal information in the detection pipeline with LSTM \cite{mccrae3d} or 3D convolution\cite{luo2018fast}. Yet, they do not achieve a compelling performance on a public dataset. 

In recent years, large-scale autonomous driving datasets such as Waymo Open Dataset\cite{sun2019scalability} provide continuous, long sequences of labeled frames. Hence, temporal information can be readily utilized for better detection performances.

In our submission, we utilize a simple yet effective way to leverage the temporal information by combining multiple frames of point clouds. This method is helpful in two ways. First, multiple consecutive point clouds create a denser resultant point cloud. Second, moving objects leave a long trail in the combined object, making heading predictions easier. We also observe a stronger model on 3D detection also performs better on domain adaptation.

\begin{figure*}[t]
\centering
  \includegraphics[width=\textwidth]{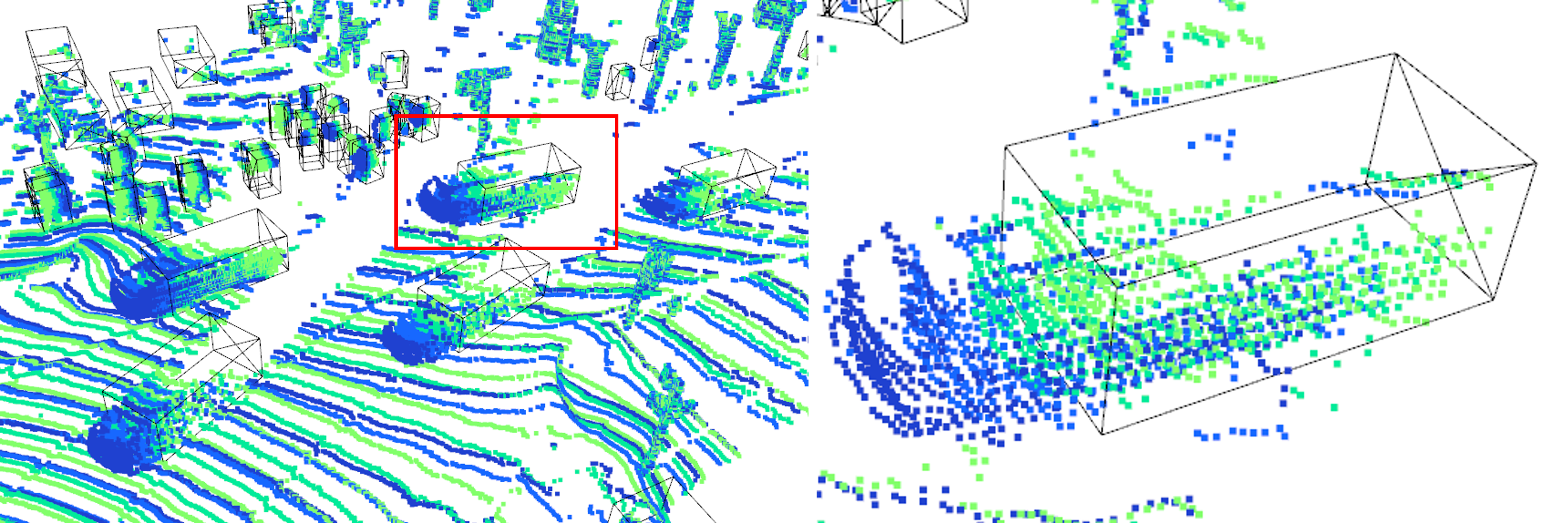}
  \caption{Fast moving objects such as cars can leave a very long trail in the concatenated point cloud. The preceding point clouds should not be included in the predicted bounding boxes. Hence, it is necessary to add the relative timestamp in the point clouds to resolve the ambiguity. Here, the point clouds are rendered the same way as that in Figure \ref{fig:temporal}}
\label{fig:long_trail}
\end{figure*}

\section{Data Conversion Toolkit}

\begin{table*}[h]
\begin{center}
\begin{tabular}{|c|c|c|c|c|c|}
\hline
Temporal & Eval Set & Vehicle & Pedestrian & Cyclist & All \\
\hline\hline
\xmark  & Test & 0.6465/0.6409 & 0.6371/0.5548  &  0.6367/0.6250 & 0.6401/0.6069 \\
\cmark  & Test & 0.6640/0.6591 & 0.6600/0.6093 & 0.6513/0.6410 & 0.6584/0.6365 \\
\xmark  & DA Test & 0.4847/0.4792 & 0.4613/0.4321  & 0.1884/0.1871 & 0.3781/0.3662 \\
\cmark  & DA Test & 0.4550/0.4477 & 0.4599/0.4312  & 0.0153/0.0148 & 0.3101/0.2979  \\
\hline
\end{tabular}
\end{center}
\caption{The effects of adding the temporal information. The results are shown in the form of mAP(L2)/mAPH(L2). Having the temporal information consistently improves model performance across test sets and object classes. More importantly, the temporal information is helpful to pedestrian's heading prediction as the gap between mAP and mAPH are effectively reduced after adding the temporal information}
\label{tab:temporal}
\end{table*}

Due to the prominent influence of KITTI on the research community, many existing projects take in KITTI-format data. Hence, we implement a data conversion toolkit\footnote{https://github.com/caizhongang/waymo\_kitti\_converter} to convert WOD data to the KITTI-format. The toolkit also converts KITTI-format prediction results back to WOD-format binary files for evaluation and submission.

\subsection{Point clouds}
Both KITTI and WOD have the LiDAR coordinate system convention being front-left-up. However, WOD has multiple LiDARs: the common approach is to transform all point clouds to the Self-Driving Car(SDC)'s coordinate system, which has the same orientation as the top LiDAR but placed at the bottom of the car.

\subsection{Images}
The conversion is trivial; the only note point is that the front camera is indexed 2 in KITTI but 0 in WOD.

\subsection{Bounding box labels}
The WOD's bounding boxes are annotated in the SDC's coordinate system, which makes it the same as the point clouds (after fusion). However, KITTI provides the bounding boxes annotations in the ``reference camera coordinates"\cite{geiger2013vision}, which does not exist in WOD. Hence, we create a virtual reference camera coordinates (see \textbf{Calibration}) and project the bounding boxes to the virtual coordinates. Note that the heading in WOD takes positive x-axis of SDC's coordinate system as 0\degree (the ``front" direction of the SDC) whereas in KITTI, it is the positive x-axis of the reference camera coordinates (the ``right" direction of the SDC)\footnote{This is different from the description in the original paper\cite{geiger2013vision}, but we take reference from \cite{kitti_object_vis} and confirm it through visualization}. In addition, length, width, and height correspond to x-, y- and z-axis of SDC's coordinates in WOD whereas they correspond to z-, x- and y-axis of reference camera coordinates in KITTI.

\subsection{Calibration}
In KITTI, to project a point $x_{sdc}$ in the SDC's coordinates to a pixel in the camera coordinates $y_{img}$, the point has to be projected to the reference camera coordinates (via $T_{ref}$), then to rectified camera coordinates (via $R_{rect}$), and finally to the image plane (via $P_{img}$):
$$y_{img} = P_{img} R_{rect} T_{ref} x_{sdc}$$
However, neither the reference camera coordinates nor the rectified camera coordinates are defined in WOD. Hence, we define the front camera's coordinates as the reference camera coordinates and $T_{rect}$ to be an identity matrix. $T_{ref}$ is taken from the extrinsic parameters of the front camera.

\subsection{Self-driving car's poses}
KITTI does provide GPU + IMU information in the raw data, however, the main training set is not continuous. WOD provides frame-wise SDC's pose information that our tool also extracts. The SDC's pose information is critical to leveraging the temporal information.

\section{Methods and Experiments}
\label{sec:methods_experiments}

We describe our methods in detail and discuss the experiment results for successful and unsuccessful attempts in this section.

\subsection{Base Model}
We adopt the 3D detection framework PCDet\cite{shi2020points}, which provides implementations of common baselines. Part A$^{2}$ is chosen as the base model for its high performance on KITTI. 

However, compared to the default hyperparameters for KITTI, we tailor them to the characteristics of WOD. By analyzing the statistics of the dataset, we set the point cloud range to be x $\in$ [-102.4, 102.4], y $\in$ [-102.4, 102.4], z $\in$ [-10, 15] to accommodate the much larger annotated area and sloppy terrains. However, we have to increase the original voxel size to 0.1 for all three dimensions due to the memory constraint, but the maximum points per voxel are increased to 10 for compensation of the reduced resolution. We also increase the upper limit of the number of voxels to be generated to 160,000 during testing.

\subsection{Temporal Information}
As the density of points is critical to detection performance, an intuitive method is to directly concatenate point clouds of multiple consecutive frames into one. This is achieved through transformation:
$$x_{a} = T_{a}^{-1} T_{i} x_{i}$$
where transformations $T$ transforms from the SDC's coordinates to the global frame. Hence, each point $x_{i}$ in the SDC's coordinates at time $i$ is transformed to $x_{a}$ in the SDC's coordinates at an anchor timestamp (the frame for evaluation). Here, we use 3 frames before the anchor frame ($i \in \{a-3, a-2, a-1\}$). For frames that have fewer than 3 preceding frames, only the available frames are used. Shown in Figure \ref{fig:temporal}, concatenation of point clouds serve two purposes. First, due to occlusion or far distance, objects can have partial or sparse point clouds. Having multiple frames provides a more complete and denser final point cloud. Second, it is challenging to determine the object's heading, especially for pedestrians which are cylindrical in shape. The concatenation of consecutive point clouds shows the trail of the pedestrian, making heading predictions easier.

However, we find empirically that the naive concatenation of point clouds does not improve performance. This is because the method is only effective for static or slow-moving objects where point clouds can be directly stacked. For fast-moving objects, however, the objects form a long trail as shown in Figure \ref{fig:long_trail}, leading to false positive bounding boxes on the preceding point clouds or bounding boxes with excessively large dimensions. Inspired by \cite{hu2019you}, we add the timestamp as an additional attribute to the point cloud, besides x, y, z, and intensity. This simple method passes in the temporal information and allows the network to understand the motion of the objects. Our hypothesis is supported by the experiment results in Table \ref{tab:temporal}.

\begin{table}[t]
\begin{center}
\begin{tabular}{|l|c|}
\hline
Model & Vehicle APH(L2)\\
\hline\hline
Pretrained (P) &  0.6236 \\
P + Remove Empty & 0.6260 \\
P + Dimension Suppression & 0.6237 \\
P + Multi-Epoch Ensemble & 0.6244 \\
\hline
\end{tabular}
\end{center}
\caption{Ensemble and postprocessing, evaluated on the validation set. The model is pretrained on the main training set. Remove Empty: remove bounding boxes with no points. Dimension Suppression: remove bounding boxes with all three dimensions $<$ 0.5 m. Multi-Epoch Ensemble: ensemble of high-performing models trained with different number of epochs}
\label{tab:tricks}
\end{table}

\subsection{Ensemble}
Since it is a common practice to use multiple models for 3D detection, we trained three Part A$^{2}$ models for vehicles, pedestrians, and cyclists respectively. The predicted bounding boxes of three models are directly combined. 

Besides the ensemble of expert models, we attempt to ensemble multiple high-performing checkpoint models from different epochs of training. However, it is shown in Table \ref{tab:tricks} that such strategy gives marginal improvement only.

\subsection{Postprocessing}
Some erroneous labels in the WOD (we find some signs are labeled as the vehicle) causes the model to predict bounding boxes with extremely small dimensions. We remove these bounding boxes that are unreasonably small (for example, all dimensions are smaller than 0.5 meters for vehicles), but since all these bounding boxes have a very low confidence score anyway, the improvement is marginal.

Since we concatenate multiple point clouds during inference, some output bounding boxes do not contain any points belonging to the current frame. Hence, we remove all these bounding boxes and observe a small improvement.

\subsection{Fine Tuning on Domain Adaptation}

After obtaining reasonably good models, we attempt to fine-tune them on the labeled data of domain adaptation. However, it is shown in Table \ref{tab:finetune} that the pretrained model performs the best. This observation shows the significant gap between the source and the target domains that cannot be naively closed through fine-tuning. 

\section{Conclusion}
We show in this report that temporal information is helpful to detection and in turn, beneficial to domain adaptation on the point clouds. However, it is noted that domain adaptation remains a largely unsolved problem that should draw the research community's attention.

{\small
\bibliographystyle{ieee_fullname}
\bibliography{references}
}

\begin{table*}[t]
\begin{center}
\begin{tabular}{|c|c|c|c|c|c|c|c|c|c|c|c|}
\hline
PT     & FB     & Eval Set & w/o FT & Epoch 1 & Epoch 2 & Epoch 3 & Epoch 4 & Epoch 5 & Epoch 6 & Epoch 7 & Epoch 8 \\
\hline\hline
\cmark & \xmark & DA Train & 0.4596 & 0.4254  & 0.4366 & 0.4462 & 0.4488 & -      & -      & -      & -      \\
\cmark & \xmark & DA Val   & 0.4010 & 0.3659  & 0.3748 & 0.3770 & 0.3757 & -      & 0.3669 & -      & 0.3778 \\
\cmark & \cmark & DA Val   & 0.4010 & 0.3244  & 0.3257 & 0.3257 & 0.3305 & -      & -      & -      & -      \\
\xmark & \xmark & DA Val   & 0.4010 & 0.0     & 0.0    & 0.0    & 0.0034 & 0.2115 & 0.2223 & 0.2003 & 0.1656 \\
\hline
\end{tabular}
\end{center}
\caption{Fine tuning for domain adaptation (DA). PT: pretrained. FB: frozen backbone. w/o FT: without fine tuning. The experiment results show that naive fine tuning is ineffective for the domain adaption}
\label{tab:finetune}
\end{table*}

\end{document}